\def\year{2018}\relax
\documentclass[letterpaper]{article} 
\usepackage{aaai18}  
\usepackage{times}  
\usepackage{helvet}  
\usepackage{courier}  
\usepackage{url}  
\usepackage{graphicx}  
\usepackage{float}
\usepackage{subfig}
\usepackage{epstopdf}
\usepackage{multirow}
\usepackage{epsfig}
\usepackage{amssymb}
\usepackage{todonotes}
\usepackage{soul, color}
\usepackage{amsmath}
\graphicspath{{./Images/}}
\begin{document}
%
\title{Unravelling Robustness of Deep Learning based Face Recognition Against Adversarial Attacks}

\begingroup
\centering{
\author{Gaurav Goswami$^{1,2}$, Nalini Ratha$^3$, Akshay Agarwal$^1$, Richa Singh$^1$, Mayank Vatsa$^1$\\
$^{1}$IIIT-Delhi, India $^{2}$IBM IRL, Bangalore, India, $^3$IBM TJ Watson Research Center, USA\\
{\tt\small \{gauravgs, akshaya, rsingh, mayank\}@iiitd.ac.in}, {\tt\small ratha@us.ibm.com} \\
}}
 \endgroup

%
\maketitle
\begin{abstract}
Deep neural network (DNN) architecture based models have high expressive power and learning capacity. However, they are essentially a black box method since it is not easy to mathematically formulate the functions that are learned within its many layers of representation. Realizing this, many researchers have started to design methods to exploit the drawbacks of deep learning based algorithms questioning their robustness and exposing their singularities. In this paper, we attempt to unravel three aspects related to the robustness of DNNs for face recognition: (i) assessing the impact of deep architectures for face recognition in terms of vulnerabilities to attacks inspired by commonly observed distortions in the real world that are well handled by shallow learning methods along with learning based adversaries; (ii) detecting the singularities by characterizing abnormal filter response behavior in the hidden layers of deep networks; and (iii) making corrections to the processing pipeline to alleviate the problem. Our experimental evaluation using multiple open-source DNN-based face recognition networks, including OpenFace and VGG-Face, and two publicly available databases (MEDS and PaSC) demonstrates that the performance of deep learning based face recognition algorithms can suffer greatly in the presence of such distortions. The proposed method is also compared with existing detection algorithms and the results show that it is able to detect the attacks with very high accuracy by suitably designing a classifier using the response of the hidden layers in the network. Finally, we present several effective countermeasures to mitigate the impact of adversarial attacks and improve the overall robustness of DNN-based face recognition.
\end{abstract}

\section{Introduction}

\begin{figure}[!t]
	\centering
   {	
	\includegraphics[width=0.45\textwidth]{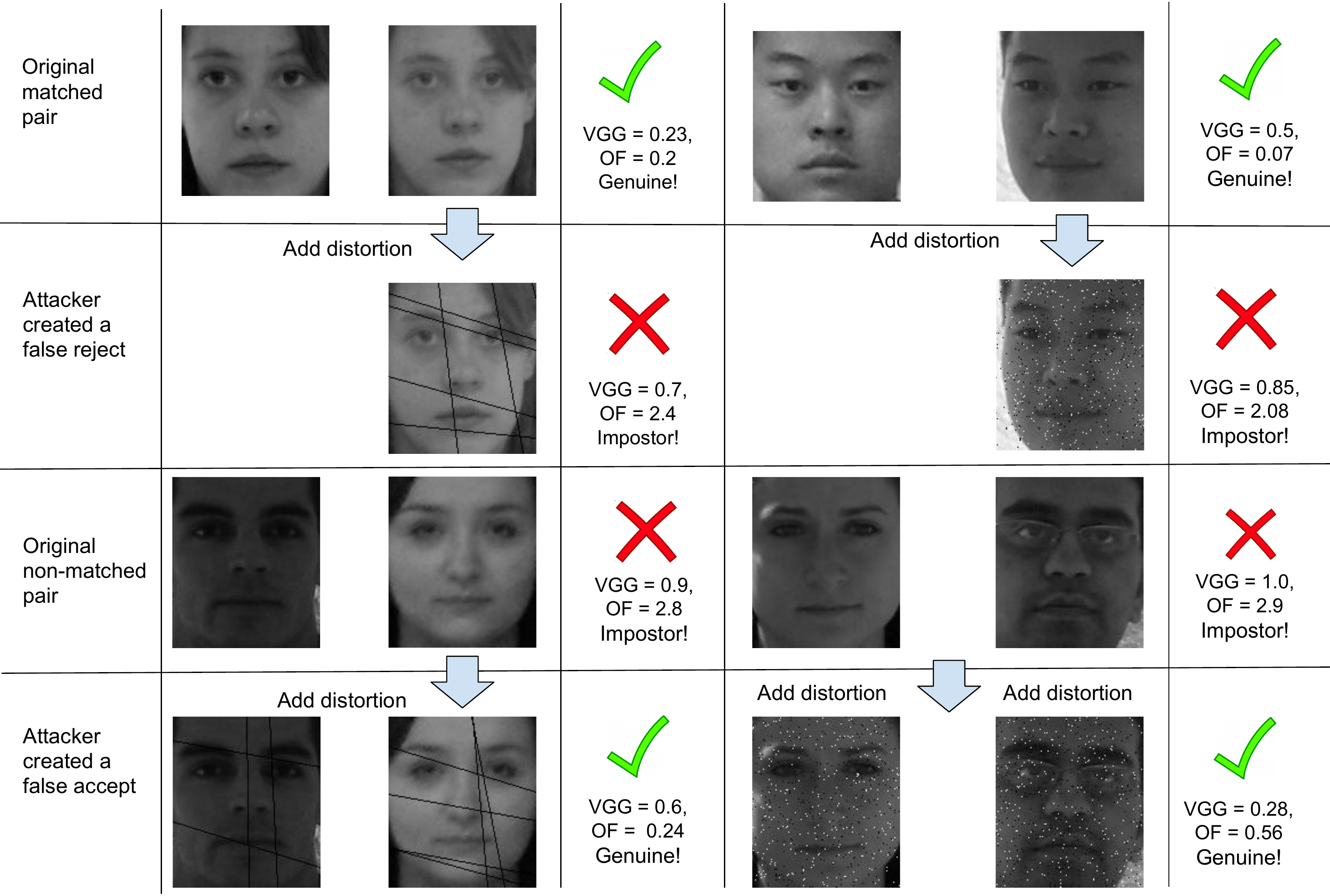}}
	\caption{\label{faceattack} We show that deep learning based OpenFace (OF) and VGG-Face can be deceived even by image processing operations that mimic real world distortions.}
\end{figure}

\begin{table*}[!t]
\centering
\caption{Literature review of adversarial attack generation and detection algorithms.}
\label{adversary}
\setkeys{Gin}{keepaspectratio}
\resizebox*{0.95\textwidth}{1.0\textheight} {
\begin{tabular}{|l|l|l|} \hline
Adversary & Authors      & Description                                                    \\ \hline
\multirow{8}{*}{Generation} & \citeauthor{szegedy2013intriguing}, \citeyear{szegedy2013intriguing}  & L-BFGS: $L(x+\rho,l)+\lambda||\rho||^2 \  s.t. \ x_i+\rho_i \in {[b_{min},b_{max}]}$ \\ \cline{2-3}
& \citeauthor{goodfellow}, \citeyear{goodfellow}       & FGSM: $x_0+\epsilon*(\triangledown_x L(x_0,l_0)$                             \\ \cline{2-3}
& \citeauthor{kurakin2016adversarial},  \citeyear{kurakin2016adversarial}      & I-FGSM: $x_{k+1}=x_k+\epsilon*(\triangledown_x L(x_0,l_0)$       \\ \cline{2-3}
& \citeauthor{papernot2016limitations},  \citeyear{papernot2016limitations} & Saliency Map: $l_0$ distance optmization \\ \cline{2-3}
& \citeauthor{moosavi2015deepfool},  \citeyear{moosavi2015deepfool} & DeepFool: $for\ each\ class, l \neq l_0, minimize\ d(l,l_0) $                         \\ \cline{2-3}
& \citeauthor{carlini2017towards},  \citeyear{carlini2017towards}   & C $\&$ W: $l_p$ distance metric optimization                                 \\ \cline{2-3}
& \citeauthor{moosavi2017universal},  \citeyear{moosavi2017universal} & Universal: Distribution based perturbation \\ \cline{2-3}
& \citeauthor{DBLP:journals/corr/RauberBB17},  \citeyear{DBLP:journals/corr/RauberBB17} & Blackbox: Uniform, Gaussian, Salt and Pepper, Gaussian Blur, Contrast \\ \hline

\multirow{7}{*}{Detection} & \citeauthor{grosse2017statistical},  \citeyear{grosse2017statistical}        & Statistical test for attack and genuine data distribution                      \\ \cline{2-3}
& \citeauthor{gong2017adversarial,metzen2017detecting}, 2017        & Neural network based classification               \\ \cline{2-3}
& \citeauthor{feinman2017detecting},  \citeyear{feinman2017detecting} & Randomized network using Dropout at both training and testing \\ \cline{2-3}
& \citeauthor{bhagoji2017dimensionality},  \citeyear{bhagoji2017dimensionality}        & PCA based dimensionality reduction algorithm  \\ \cline{2-3}
& \citeauthor{DBLP:journals/corr/LiangLSLSW17},  \citeyear{DBLP:journals/corr/LiangLSLSW17} & Quantization and smoothing based image processing \\ \cline{2-3}
& \citeauthor{lu2017safetynet},  \citeyear{lu2017safetynet} & Quantize ReLU output for discrete code + RBF SVM  \\ \cline{2-3}
& \citeauthor{das2017keeping},  \citeyear{das2017keeping} & JPEG compression to reduce the effect of adversary \\ \hline
\end{tabular}}
\end{table*}


Deep \textit{learning} paradigm has seen significant proliferation in face recognition due to the convenience of obtaining large training data, availability of inexpensive computing power and memory, and utilization of cameras at multiple places. Several algorithms such as DeepFace \cite{deepface}, DeepID \cite{deepid}, FaceNet \cite{facenet}, and Liu \textit{et al.} \shortcite{baidu} are successful examples of the coalesce of deep learning and face recognition. However, it is also known that machine learning algorithms are susceptible to \textit{adversaries} which can cause the classifier to yield incorrect results. Most of the time these adversaries are unintentional and are in the form of outliers. Recently, it has been shown that \textit{fooling images} can be generated in such a manner where humans can correctly classify the images but deep learning algorithms misclassify them \cite{goodfellow}, \cite{Nguyen}. As shown in Table \ref{adversary}, such images can be generated via evolutionary algorithms \cite{Nguyen} or adversarial sample crafting using the fast gradient sign method \cite{goodfellow}. Sharif et al. \shortcite{ccs} explored threat models by creating \textit{perturbed eye-glasses} to fool face recognition algorithms. An adversarial attack on face recognition is not acceptable as face biometric gets used in many high security applications such as passports, visa, and other law enforcement documents.

It is our assertion that it is not required to attack the system with sophisticated learning based attacks; and attacks such as adding random noise or horizontal and vertical black grid lines in the face image cause reduction in face verification accuracies. Samples images in Figure \ref{faceattack} show a glimpse of the effect of image processing operations on two state-of-the-art deep learning based face recognition algorithms. To the best of our knowledge, this is the first reported research on finding singularities in deep learning based face recognition engines along with detection and mitigation of such attacks. We believe that being able to not only automatically detect but also correct adversarial samples at runtime is a crucial ability for a deep network that is deployed for real world applications. With this research, we aim to present a new perspective on potential attacks as well as a different methodology to limit their performance impact beyond simply including adversarial samples in the training data.


The objective of this paper is three-fold: (i) We demonstrate that the performance of deep learning based face recognition algorithms can be significantly affected due to adversarial attacks - both image processing based adversarial attacks and adversarial samples generated in context to the recognition architecture. (ii) The first key step in taking countermeasures against such adversarial attacks is to be able to reliably determine which images contain such distortions. We propose and evaluate a methodology for automatic detection of such attacks using the response from hidden layers of the DNN. (iii) Once identified, the distorted images may be rejected for further processing or rectified using appropriate preprocessing techniques to prevent degradation in performance. To address this challenge without increasing the failure to process rate (by rejecting the samples), the third contribution of this research is a novel technique of selective dropout in the DNN to mitigate these adversarial attacks. While we have showcased results with multiple deep face networks in this paper, we have used VGG to report the detection and mitigation results for \textit{DeepFool} and \textit{Universal} adversarial perturbations since it is the only network for which the authors have provided pre-computed models.

\section{Adversarial Attacks on Deep Learning based Face Recognition}

    In this section, we discuss the proposed adversarial distortions that are able to degrade the performance of deep learning face recognition algorithms.
    Let $\mathbf{x}$ be the input to a deep learning based face recognition algorithm and $l$ be the output class label (in case of identification, it is an identity label and for verification, it is \textit{same} or \textit{different}). An adversarial attack function $a(\cdot)$, when applied to the input face image, falsely changes the predicted identity label. In other words, if $a(\mathbf{x}) = l'$ where, $l \neq l'$, then $a$ is a successful adversarial attack on the network. While adversarial learning has been used in literature to showcase that the function $a(\cdot)$ can be obtained via optimization based on network gradients, in this research, we explore a different approach. We evaluate the robustness of deep learning based face recognition in the presence of image processing based distortions. Based on the information required in their design, these distortions can be considered at image-level or face-level. We propose two image-level distortions: (a) grid based occlusion, and (b) most significant bit based noise, along with three face-level distortions: (a) forehead and brow occlusion, (b) eye region occlusion, and (c) beard-like occlusion. 

\subsection{Image-level Distortions}

\begin{figure}[t]
\centering
{\includegraphics[scale=0.37]{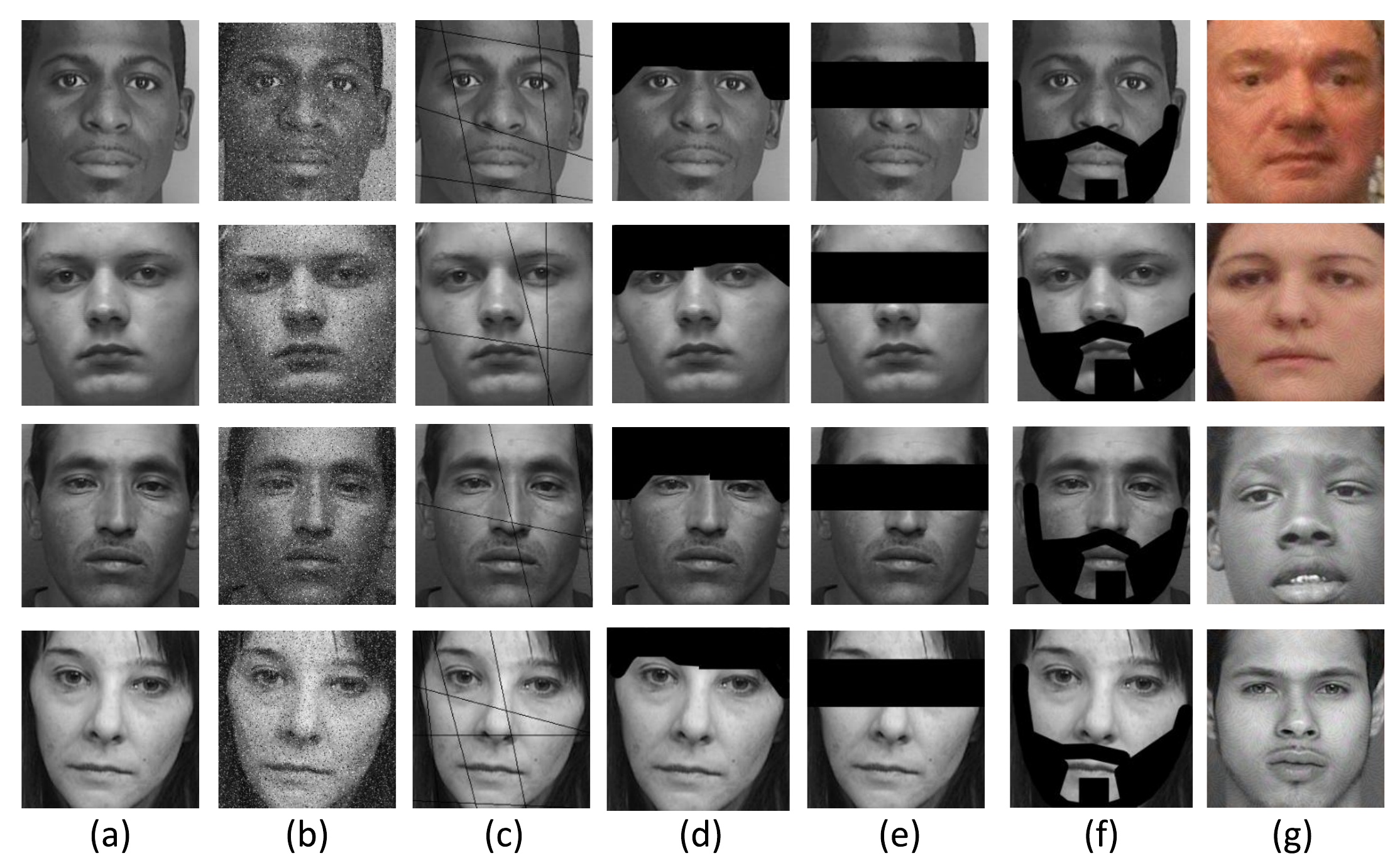}}
\caption{\label{distortions} Sample images representing the (b) grid based occlusion (Grids), (c) most significant bit based noise (xMSB), (d) forehead and brow occlusion (FHBO), (e) eye region occlusion (ERO), and (f) beard-like occlusion \cite{tejasdisguise} (Beard) distortions when applied to the (a) original images. (g) is the Universal perturbed \cite{moosavi2017universal} images of PaSC and MEDS databases.}
\end{figure}

Distortions that are not specific to faces and can be applied to an image of any object are categorized as image-level distortions. In this research, we have utilized two such distortions, grid based occlusion and most significant bit change based noise addition.  Figure \ref{distortions}(b) and \ref{distortions}(c) present sample outputs of image-level distortions.

\subsubsection{Grid based Occlusion}

For the grid based occlusion (termed as Grids) distortion, we select a number of points $P = \{p_1, p_2, ..., p_n\}$ along the upper (${y=0}$) and left (${x=0}$) boundaries of the image according to a parameter $\rho_{grids}$. The parameter $\rho_{grids}$ determines the number of grids that are used to distort each image with higher values resulting in a denser grid, i.e., more grid lines. For each point $p_i = (x_i, y_i)$, we select a point on the opposite boundary of the image, $p'_i = (x'_i, y'_i)$, with the condition if $y_i = 0$, then $y'_i = H$ and if $x_i = 0$ then $x'_i = W$, where, $W \times H$ is the size of the input image. Once a set of pairs corresponding to points $P$ and $P'$ have been selected for the image, one pixel wide line segments are created to connect each pair, and each pixel lying on these lines is set to $0$ grayscale value.

\subsubsection{Most Significant Bit based Noise}

For the most significant bit based noise (xMSB) distortion, we select three sets of pixels $\mathcal{X}_1, \mathcal{X}_2, \mathcal{X}_3$ from the image stochastically such that $\vert \mathcal{X}_i \vert = \phi_i \times W \times H$, where $W \times H$ is the size of the input image. The parameter $\phi_i$ denotes the fraction of pixels where the $i^{th}$ most significant bit is flipped. The higher the value of $\phi_i$, the more pixels are distorted in the $i^{th}$ most significant bit. For each $\mathcal{P}_j \in X_i, \forall i \in [1,3]$, we perform the following operation:

\begin{equation}
\mathcal{P}_{kj} = \mathcal{P}_{kj} \oplus 1
\end{equation}

\noindent{where,} $\mathcal{P}_{kj}$ denotes the $k^{th}$ most significant bit of the $j^{th}$ pixel in the set and $\oplus$ denotes the bitwise XOR operation. It is to be noted that the sets $\mathcal{X}_i$ are not mutually exclusive and may overlap. Therefore, the total number of pixels affected by the noise is at most $|\mathcal{X}_1 + \mathcal{X}_2 + \mathcal{X}_3|$ but may also be lower depending on the stochastic selection.

\subsection{Face-level Distortions}


Face-level distortions specifically require face-specific information, e.g. location of facial landmarks. The three face-level region based occlusion distortions are applied after performing automatic face and facial landmark detection. In this research, we have utilized the open source DLIB library \cite{dlib09} to obtain the facial landmarks. Once facial landmarks are identified, they are used along with their boundaries for masking. To obscure the eye region, a singular occlusion band is drawn on the face image as follows:

\begin{equation}
I\{x,y\} = 0, \forall x \in [0,W], y \in \Big[y_e - \frac{d_{eye}}{\psi}, y_e + \frac{d_{eye}}{\psi}\Big]
\end{equation}

\noindent{H}ere, $y_e = \left(\frac{y_{le}+y_{re}}{2}\right)$, and $(x_{le}, y_{le})$ and $(x_{re}, y_{re})$ are the locations of the left eye center and the right eye center, respectively. The inter-eye distance $d_{eye}$ is calculated as: $x_{re} - x_{le}$ and $\psi$ is a parameter that determines the width of the occlusion band. Similar to the eye region occlusion (ERO), the forehead and brow occlusion (FHBO) is created where facial landmarks on forehead and brow regions are used to create a mask. For the beard-like occlusion, outer facial landmarks along with nose and mouth coordinates are utilized to create the mask as combinations of individually occluded regions. Figure \ref{distortions} (d), (e), and (f) illustrate the samples of face-level distortions.

%

\subsection{Learning based Adversaries}
Along with the proposed image-level and face-level distortions, we also analyze the effect of adversarial samples generated using two existing adversarial models: DeepFool \cite{moosavi2015deepfool} and Universal Adversarial Perturbations \cite{moosavi2017universal}.

\begin{table}[!t]
\centering
\caption{Characteristics of the databases used for adversarial attack generation and detection.}
\label{DB-stats}
\begin{tabular}{|l|l|l|} \hline
Database & Subjects & Images  \\ \hline
PaSC \cite{pasc}    & 293      & 4,688     \\ \hline
MEDS-II \cite{meds}  & 518      & 858     \\ \hline
\end{tabular}
\end{table}

\begin{table*}[!t]
\centering
\caption{\label{tab:distortions} Verification performance of existing face recognition algorithms in the presence of different distortions on the PaSC and MEDS databases. All values indicate genuine accept rate (\%) at 1\% false accept rate.}
\setkeys{Gin}{keepaspectratio}
\resizebox*{0.85\textwidth}{0.95\textheight} {
\begin{tabular}{|l||c|c|c|c|c|c||c|c|c|c|c|c|}
\hline
\multirow{2}{*}{System} & \multicolumn{6}{c}{MEDS} & \multicolumn{6}{c|}{PaSC}\tabularnewline
\cline{2-13}
 & Original & Grids & xMSB & FHBO & ERO & Beard & Original & Grids & xMSB & FHBO & ERO & Beard\tabularnewline
\hline
COTS & 24.1 & 20.9 & 14.5 & 19.0 & 0.0 & 24.8 & 40.3 & 24.3 & 19.1 & 13.0 & 0 & 6.2\tabularnewline
\hline
OpenFace & 66.7 & 49.5 & 43.8 & 47.9 & 16.4 & 48.2 & 39.4 & 10.1 & 10.1 & 14.9 & 6.5 & 22.6\tabularnewline
\hline
VGG-Face & 78.4 & 50.3 & 45.0 & 25.7 & 10.9 & 47.7 & 54.3 & 3.2 & 1.3 & 15.2 & 8.8 & 24.0\tabularnewline
\hline
LightCNN & 89.3 & 80.1 & 71.5 & 62.8 & 26.7 & 70.7 & 60.1 & 24.6 & 29.5 & 31.9 & 24.4 & 38.1\tabularnewline
\hline
L-CSSE & 89.1 & 81.9 & 83.4 & 55.8 & 27.3 & 70.5 & 61.2 & 43.1 & 36.9 & 29.4 & 39.1 & 39.8\tabularnewline
\hline
\end{tabular}
}
\end{table*}

\section{Adversarial Distortions: Results and Analysis}\label{sec:exp}
In this section, we first provide a brief overview of the deep face recognition networks, databases, and respective experimental protocols that are used to conduct the face verification evaluations. We attempt to assess how the deep networks perform in the presence of different kinds of proposed distortions to emphasize the need for addressing such attacks.

\subsection{Databases}

We use two publicly available face databases for our experiments, namely, the Point and Shoot Challenge (PaSC) database \cite{pasc} and the Multiple Encounters Dataset (MEDS) \cite{meds}. The PaSC database contains still-to-still and video-to-video matching protocols. We use the frontal subset of the still-to-still protocol which contains 4,688 images pertaining to 293 individuals which are divided into equal size target and query sets. Each image in the target set is matched to each image in the query set and the resulting $2344 \times 2344$ score matrix is used to determine the verification performance. 

The MEDS-II database contains a total of 1,309 faces pertaining to 518 individuals. Similar to the case of PaSC, we utilize the metadata provided with the MEDS release 2 database to obtain a subset of 858 frontal face images from the database. Each of these images is matched to every other image and the resulting $858 \times 858$ score matrix is utilized to evaluate the verification performance.
For evaluating performance under the effect of distortions, we randomly select 50\% of the total images from each database and corrupt them with the proposed distortions separately. These distorted sets of images are utilized to compute the new score matrices for each case.

\subsection{Existing Networks and Systems}

In this research, we utilize the OpenFace (\citeauthor{openface}), VGG-Face \cite{parkhi2015deep}, LightCNN \cite{wulight}, and L-CSSE \cite{csse} networks to gauge the performance of deep face recognition algorithms in the presence of the aforementioned distortions. The OpenFace library is an open source implementation of FaceNet \cite{facenet} and is openly available to all members of the research community for modification and experimental usage. The VGG deep face network is a deep convolutional neural network (CNN) with 11 convolutional blocks where each convolution layer is followed by non-linearities such as ReLU and max pooling. LightCNN is another publicly available deep network architecture for face recognition that is a CNN with maxout activations in each convolutional layer and achieves good results with just five convolutional layers. L-CSSE is a supervised autoencoder formulation that utilizes a class sparsity based supervision penalty in the loss function to improve the classification capabilities of autoencoder based deep networks. In order to assess the relative performance of deep face recognition with a non-deep learning based approach, we compare the performance of these deep learning based algorithms with a commercial-off-the-shelf (COTS) matcher. No fine-tuning is performed for any of these algorithms before evaluating their performance on the test databases.



\subsection{Results and Analysis}


Table \ref{tab:distortions} summarizes the effect of image processing based adversarial distortions on OpenFace, VGG-Face, LightCNN, L-CSSE, and COTS. On the PaSC database, as shown in Table \ref{tab:distortions}, while OpenFace and COTS perform comparably to each other at about 1\% false accept rate (FAR), OpenFace performs better than the COTS algorithm at all further operating points when no distortions are present. However, we observe a sharp drop in OpenFace performance when any distortion is introduced in the data. For instance, with grids attack, at 1\% FAR, the GAR of OpenFace drops by 29.3\% and of VGG by 28.1\%, whereas the performance of  COTS only drops by 16\% which is about half the drop compared to what OpenFace and VGG-Face experience. We notice a similar scenario in the presence of noise attack where the performance of OpenFace and VGG drops down by about 29\% as opposed to the loss of 21.2\% observed by COTS. In cases of LightCNN and L-CSSE, they both have shown higher performance with original images; however, as shown in Table \ref{tab:distortions}, similar level of drops are observed. It is to be noted that for xMSB and grid attack, L-CSSE is able to achieve relatively better performance because L-CSSE is a supervised version of autoencoder which can handle \textit{noise} better. Overall, deep learning based algorithms experience higher performance drop as opposed to the non-deep learning based COTS. In the case of occlusions, however, deep learning based algorithms suffer less as compared to COTS. It is our assessment that the COTS algorithm fails to perform accurate recognition with the highly limited facial region available in the low-resolution PaSC images in the presence of occlusions. Similar performance trends are observed on the MEDS database on which for original images, deep learning based algorithms outperform the COTS matcher with a GAR of 60-89\% at 1\% FAR respectively as opposed to 24.1\% by COTS. The accuracy of deep learning algorithms drops significantly more than the accuracy of COTS.

\begin{figure}
\begin{center}
   \includegraphics[width=0.6\linewidth]{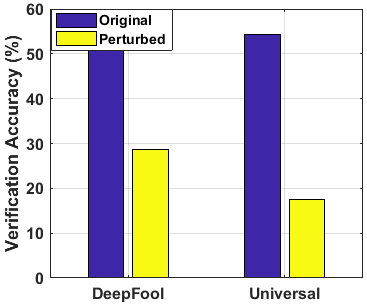}
\end{center}
\caption{Bar graph showing the effect of perturbation on the VGG-Face model. Verification accuracy is reported at 1\% GAR.}
\label{fig:bar-PaSC-adversary}
\end{figure}

We next performed a similar analysis with learning based adversaries on the PaSC database. The results of VGGFace model with original and perturbed images are shown in Figure \ref{fig:bar-PaSC-adversary}. It is interesting to observe that the drop in accuracy obtained by simple image processing operations is equivalent to the reduction achieved by learned adversaries. This clearly shows that deep models are not resilient to even simple perturbations and therefore, it is very important to devise effective strategies for detection and mitigation of attacks.


\section{Detection and Mitigation of Adversarial Attacks}

As we can see in the previous section, adversarial attacks can substantially reduce the performance of usually accurate deep neural network based face recognition methods. Therefore, it is essential to address such singularities in order to make face recognition algorithms more robust and useful in real world applications. In this section, we propose novel methodologies for detecting and mitigating adversarial attacks. First, we provide a brief overview of a deep network followed by the proposed algorithms and their corresponding results.

\begin{figure}
\resizebox{\linewidth}{!}{
\includegraphics{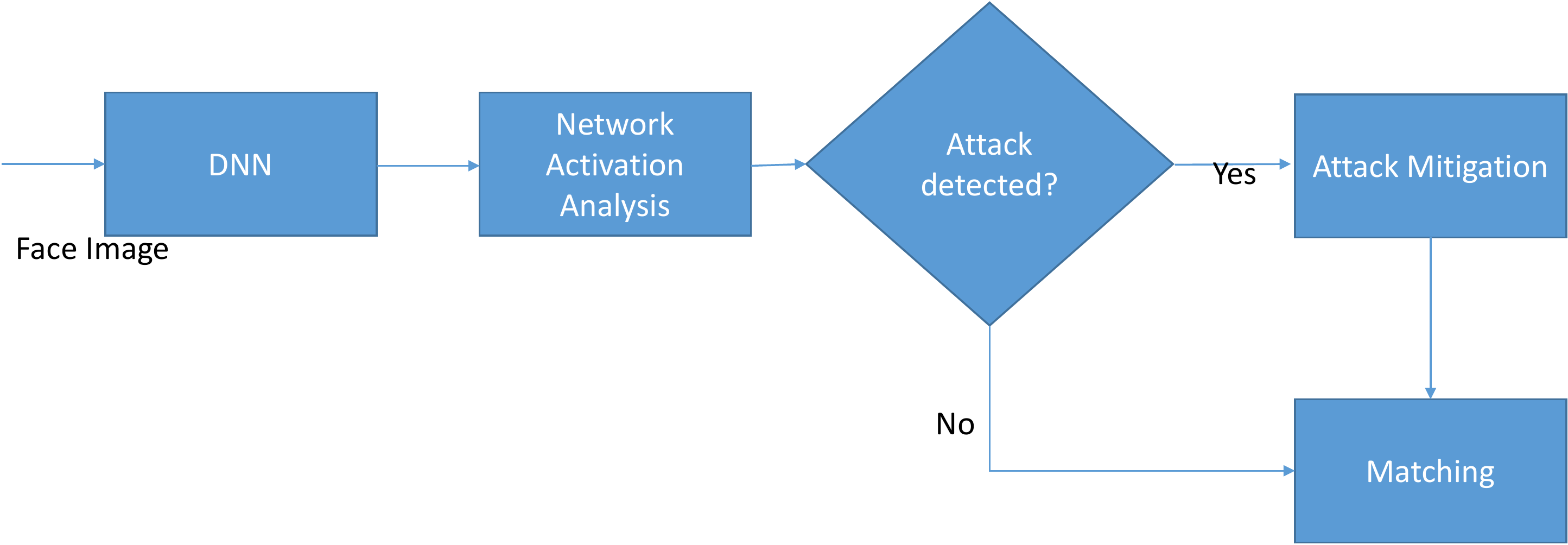}}
\caption{\label{fig:flowdiag} Flow chart for the proposed detection and mitigation methodology.} 
\end{figure}

Each layer in a deep neural network essentially learns a function or representation of the input data. The final feature computed by a deep network is derived from all of the intermediate representations in the hidden layers. In an ideal scenario, the internal representation at any given layer for an input image should not change drastically with minor changes to the input image. However, that is not the case in practice as proven by the existence of adversarial examples. The final features obtained for a distorted and undistorted image are measurably different from one another since these features map to different classes. Therefore, it is implied that the intermediate representations also vary for such cases. It is
our assertion that the internal representations computed at each layer are different for distorted images as compared to undistorted images. Therefore, in order to detect whether an incoming image is perturbed in an adversarial manner, we decide that it is distorted if its layer-wise internal representations deviate substantially from the corresponding mean representations. The overall flow of the detection and mitigation algorithms is summarized in Figure \ref{fig:flowdiag}.

\subsection{Network Analysis and Detection}


In order to develop adversarial attack detection mechanism, we first analyze the filter responses in CNN architecture. Network visualization analysis showcases the filter responses for a distorted image at selected intermediate layers that demonstrate the most sensitivity towards noisy data. We can see that many of the filter outputs primarily encode the noise instead of the input signal. We observe that the deep network based representation is more sensitive to the input and while that sensitivity results in a more expressive representation that offers higher performance in case of undistorted data, it also compromises the robustness towards noise such as the proposed distortions. Since each layer in a deep network learns increasingly more complicated functions of the input data based on the functions learned by the previous layer, any noise in the input data is also encoded in the features thus leading to a higher reduction in the discriminative capacity of the final learned representation. Similar conclusions can also be drawn from the results of other existing adversarial attacks on deep networks, where the addition of a noise pattern leads to spurious classification \cite{goodfellow}.

To counteract the impact of such attacks and ensure practical applicability of deep face recognition, the networks must either be made more robust towards noise at a layer level during training or it must be ensured that any input is preprocessed to filter out any such distortion prior to computing its deep representation for recognition.

In order to detect distortions we compare the pattern of the intermediate representations for undistorted images with distorted images at each layer. The differences in these patterns are used to train a classifier that can categorize an unseen input as an undistorted/distorted image. In this research, we use the VGG-Face \cite{parkhi2015deep} and LightCNN \cite{wulight} networks to devise and evaluate our detection methodology. From the 50,248 frontal face images in the CMU Multi-PIE database \cite{gross2010multi}, 40,000 are randomly selected and used to compute a set of layer-wise mean representations, $\mathcal{\mu}$, as follows:

\begin{equation}
\mu_i = \frac{1}{N_{train}} \Sigma_{j=1}^{N_{train}} \phi_i\left(I_j\right)
\end{equation}

\noindent{where}, $I_j$ is the $j^{th}$ image in the training set, $N_{train}$ is the total number of training images, $\mu_i$ is the mean representation for the $i^{th}$ layer of the network, and $\phi_i(I_j)$ denotes the representation obtained at the $i^{th}$ layer of the network when $I_j$ is the input. Once $\mathbf{\mu}$ is computed, the intermediate representations computed for an arbitrary image $I$ can be compared with the layer-wise means as follows:

\begin{equation}
\Psi_i(I, \mathbf{\mu}) = \Sigma_{z}^{\lambda_i} \frac{|\phi_i(I)_z - \mu_{iz}|}{|\phi_i(I)_z| + |\mu_{iz}|}
\end{equation}

\noindent{where}, $\Psi_i(I,\mathbf{\mu})$ denotes the Canberra distance between $\phi_i(I)$ and $\mu_i$, $\lambda_i$ denotes the length of the feature representation computed at the $i^{th}$ layer of the network, and $\mu_{iz}$ denotes the $z^{th}$ element of $\mu_i$. If the number of intermediate layers in the network is $N_{layers}$, we obtain $N_{layers}$ distances for each image $I$. These distances are used as features to train a Support Vector Machine (SVM) \cite{SVM} for two-class classification.

\subsection{Mitigation: Selective Dropout}

An ideal automated solution should not only automatically detect but also mitigate the effect of an adversarial attack so as to maintain as high performance as possible. Therefore, the next step in defending against adversarial attack is mitigation.  This can be achieved by discarding or preprocessing (e.g. denoising) the affected regions. In order to accomplish these objectives, we again utilize the characteristics of the output produced in the intermediate layers of the network. We select 10,000 images from the Multi-PIE database that are partitioned into 5 mutually exclusive and exhaustive subsets of 2,000 images each. Each subset is processed using a different distortion. The set of 10,000 distorted images thus obtained contains 2,000 images pertaining to each of the five proposed distortions. We use a smaller separate Multi-PIE subset of 1,680 faces (5 per subject) for training the algorithm on DeepFool and Universal perturbations. Using this data, we compute a filter-wise score per layer that estimates the particular filter's sensitivity towards distortion as follows:

\begin{equation}
\epsilon_{ij} = \Sigma_{k=1}^{N_{dis}} \|\phi_{ij}(I_{k}) - \phi_{ij}(I_{k}^{'})\|
\end{equation}

\noindent{where}, $N_{dis}$ is the number of distorted images in the training set, $\epsilon_{ij}$ denotes the score and $\phi_{ij}(\cdot)$ denotes the response of the $j^{th}$ filter in the $i^{th}$ layer, $I_k$ is the $k^{th}$ distorted image in the dataset, and $I_{k}^{'}$ is the undistorted version of $I_{k}$. Once these values are computed, the top $\eta$ layers are selected based on the aggregated $\epsilon$ values for each layer. These are the layers identified to contain the most filters that are adversely affected by the distortions in data. For each of the selected $\eta$ layers, the top $\kappa$ fraction of affected filters are disabled by modifying the weights pertaining to $0$ before computing the features. We also apply a median filter of size $5 \times 5$ for denoising the image before extracting the features. We term this approach as \textit{selective dropout}. It is aimed at increasing the network's robustness towards noisy data by removing the most problematic filters from the pipeline. We determine the values of parameters $\eta$ and $\kappa$ via grid search optimization on the training data with verification performance as the criterion.

\subsection{Experimental Protocol}

For training the detection model, we use the remaining 10,000 frontal face images from the CMU Multi-PIE database as undistorted samples. We generate 10,000 distorted samples using all five distortions with 2,000 images per distortion that are also randomly selected from the CMU Multi-PIE database. We use the same training data for universal perturbations with 10,000 distorted and 10,000 undistorted samples. For DeepFool, we use a subset of 1,680 face images from the CMU Multi-PIE database with 5 images from each of the 336 subjects with both distorted and undistorted versions for training the detection algorithm. Since the VGG-Face network has 20 intermediate layers, we obtain a feature vector of size $20$ distances for each image. We perform a grid search based parameter optimization using the $20,000 \times 20$ training matrix to optimize and learn the SVM model. For DeepFool, the size of the training data is $3,360 \times 20$. Once the model is learned, any given test image is characterized by the distance vector and processed by the SVM. The score given by the model for the image to belong to the distorted class is used as a distance metric. We observe that the metric thus obtained is able to classify distorted images on unseen databases. The mitigation algorithm is evaluated with both LightCNN and VGG-Face networks on both the PaSC and MEDS databases with the same experimental protocol as used in obtaining the verification results. 

\subsection{Results and Analysis}

\begin{table*}[!]
\centering
\caption{\label{tab:detection} Performance (accuracy \%) of the proposed detection methodology (using LightCNN and VGG-Face as the target networks) compared to two existing detection algorithms. Grids = grid based occlusion, xMSB = most significant bit based noise, FHBO = forehead and brow occlusion, ERO = eye region occlusion, and Beard = beard like occlusion.}
\resizebox{\linewidth}{!}{
\begin{tabular}{|c|c|c|c|c|c|c|c|c|}
\hline
Distortion & \multicolumn{4}{c|}{MEDS} & \multicolumn{4}{c|}{PaSC}\tabularnewline
\hline
\hline
 &  LightCNN & VGG & \cite{DBLP:journals/corr/LiangLSLSW17} & \cite{feinman2017detecting}  & LightCNN & VGG & \cite{DBLP:journals/corr/LiangLSLSW17} & \cite{feinman2017detecting}\tabularnewline
\hline
Beard  & \textbf{92.2} & 86.8 & 81.2 & 80.9 &  89.5 & \textbf{99.8} & 83.4 & 85.1\tabularnewline
\hline
ERO &  \textbf{91.9} & 86.0 & 80.4 & 80.0 &  90.6 & \textbf{99.7} & 84.9 & 84.6\tabularnewline
\hline
FHBO &  \textbf{92.9} & 84.4 & 79.8 & 79.6 &  81.7 & \textbf{99.8} & 78.3 & 77.8\tabularnewline
\hline
Grids &  68.4 & \textbf{84.4} & 62.1 & 62.4 &  89.7 & \textbf{99.9} & 85.1 & 85.7\tabularnewline
\hline
xMSB &  \textbf{92.9} & 85.4 & 80.2 & 80.9 &  93.2 & \textbf{99.8} & 88.2 & 87.9\tabularnewline
\hline
\end{tabular}}
\end{table*}

First, we present the results of the proposed algorithm in detecting whether an image contains adversarial distortions or not using the VGG and LightCNN networks. We choose these two as the model definition and weights are publicly available. Table \ref{tab:detection} presents the results of adversarial attack detection. Each distortion based subset comprises of a 50\% split of distorted and undistorted faces. These are the same sets that have been used for evaluating the performance of the three face recognition systems. As mentioned previously, the model is trained on a separate database which does not have any overlap with the test set.

The proposed detection algorithm performs almost perfectly for the PaSC database with the VGG network and maintains accuracies of 80-90\% with the LightCNN network. The lowest performance is observed on the MEDS database (classification accuracy of 68.4\% with the LightCNN network). The lower accuracies with the LightCNN can be attributed to the smaller network depth which results in smaller size features to be utilized by the detection algorithm. It is to be noted that the proposed algorithm maintains high true positive rates even at very low false positive rates across all distortions on both databases which is desirable when the cost of accepting a distorted image is much higher than a false reject for the system.
Besides exceptionally poor quality images that are naturally quite distorted, we observe that high or low illumination results in false rejects by the algorithm, i.e., falsely detected as distorted. This shows the scope of further improvement and refinement in the detection methodology. This is also another reason for lower performance with the MEDS database which has more extreme illumination cases as compared to PaSC.
We also test using the Viola Jones face detector \cite{viola2004robust} and find that, on average, approximately 60\% of the distorted faces pass face detection. Therefore, the distorted face images cannot be differentiated from undistorted faces on the basis of failing face detection. We attempt to reduce the feature dimensionality to deduce the most important features using sequential feature selection based on classification loss by an SVM model learned on a given subset of features. For the VGG-Face based model, using just the top 6 features for detection, we obtain an average accuracy of 81.7\% on MEDS and 96.9\% on PaSC database across all distortions. If we use only one most discriminative feature to perform detection, we obtain 79.3\% accuracy on MEDS and 95.8\% on PaSC on average across all distortions. This signifies that comparing the representations computed by the network in its intermediate layers indeed produces a good indicator of the existence of distortions in a given image.

\begin{figure}[!t]
\begin{center}
   \includegraphics[width=0.5\linewidth]{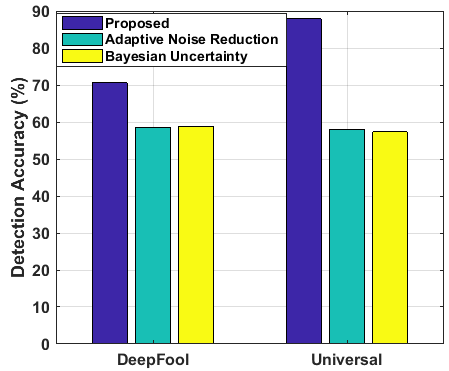}
   \includegraphics[width=0.45\linewidth]{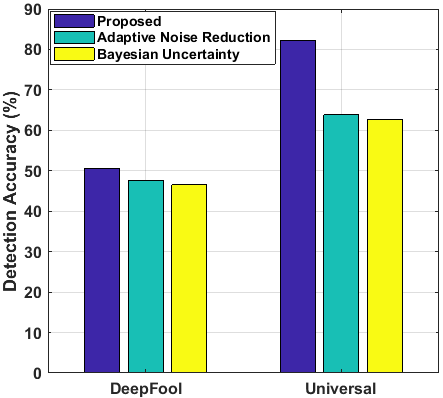}
\end{center}
\caption{Summarizing the results of the proposed and existing detection algorithms on the PaSC (Left) and MEDS (Right) databases.}
\label{fig:bar-PaSC-detection}
\end{figure}

In addition to the proposed adversarial attacks, we have also evaluated the efficacy of the proposed detection methodology on two existing attacks that utilize network architecture information for adversarial perturbation generation, i.e., DeepFool \cite{moosavi2015deepfool} and Universal adversarial perturbations \cite{moosavi2017universal}. We have also compared the performance of the proposed detection algorithm with two recent adversarial detection techniques based on adaptive noise reduction \cite{DBLP:journals/corr/LiangLSLSW17} and Bayesian uncertainty \cite{feinman2017detecting}. Same training data and protocol was used to train and test all three detection approaches. The results of detection are presented in Table \ref{tab:detection} and Figure \ref{fig:bar-PaSC-detection}. We observe that the proposed methodology is at least 11\% better at detecting DNN architecture based adversarial attacks as compared to the existing algorithms for all cases except for detecting DeepFool perturbed images from the MEDS database where it still outperforms the other approaches by more than 3\%. We believe that this is due to the fact that MEDS has overall higher image quality as compared to PaSC and even the impact of these near imperceptible perturbations (DeepFool and Universal) on verification performance is minimal for the database. Therefore, it is harder to distinguish original images from perturbed images for these distortions for all the tested detection algorithms.





Table \ref{tab:mitigation} present the results of the mitigation algorithm. Mitigation is a two-step process to enable better performance and computational efficiency. Figure \ref{fig:bar-PaSC-adversary} shows the effect of DeepFool and Universal adversary on the verification performance using VGG-Face model. First, using the proposed detection algorithm we perform selective mitigation of only those images that are considered adversarial by the learned model. Face verification results after applying the proposed mitigation algorithm on the MEDS and PaSC databases are presented in Table \ref{tab:mitigation}. We can observe that the mitigation model is able to improve the verification performance on both the databases with either network and bring it closer to the original. Thus, we see that even discarding a certain fraction of the intermediate network output, that is the most affected by adversarial distortions, results in better recognition than incorporating them into the obtained feature vector.

\begin{table}
\centering
\caption{\label{tab:mitigation} Mitigation Results (GAR (\%) at 1\% FAR) on the MEDS and PaSC databases.}
\setkeys{Gin}{keepaspectratio}
\resizebox*{0.46\textwidth}{0.46\textheight}
{
\begin{tabular}{|c|c|c|c|c|}
\hline
Algorithm & Database  & Original & Distorted & Corrected \tabularnewline \hline
\multirow{2}{*}{LCNN} & PaSC & 60.5 & 25.9 & \textbf{36.2} \\\cline{2-5}
& MEDS & 89.3 & 41.6 & \textbf{61.3} \tabularnewline
\hline

\multirow{2}{*}{VGGFace} & PaSC & 54.3 & 14.6 & \textbf{24.8}\\\cline{2-5}
& MEDS & 78.4 & 30.5 & \textbf{40.6}\tabularnewline
\hline

\end{tabular}
}
\end{table}

\section{Conclusion and Future Research Directions}

To summarize, our work has three main contributions: (i) a framework to evaluate robustness of deep learning based face recognition engines, (ii) a scheme to detect adversarial attacks on the system; and (iii) methods to mitigate adversarial attacks when detected. Playing the role of an expert level adversary, we propose five classes of image distortions in the evaluation experiment.  Using an open source implementation of Facenet, i.e., OpenFace, and the recently proposed VGG-Face, LightCNN, and L-CSSE networks, we conduct a series of experiments on the publicly available PaSC and MEDS databases. We observe a substantial loss in the performance of the deep learning based systems when compared with a non-deep learning based COTS matcher for the same evaluation data. In order to detect the attacks, we propose a network activation analysis based method in the hidden layers of the network. When an attack is reported by this stage, we invoke mitigation methods described in the paper to show that we can recover from the attacks in many situations. In the future, we will build more complex mitigation frameworks to restore to a normal level of performance. It is our assertion that with these findings, future research can be aimed at correcting such adversarial samples and incorporating various other kinds of countermeasures in deep neural networks to further increase their robustness.

\section{Acknowledgements}
Goswami was partly supported through IBM PhD Fellowship, Agarwal is partly supported by Visvesvaraya PhD Fellowship, and Vatsa and Singh are partly supported through CAI@IIIT-Delhi.

\bibliographystyle{aaai}
\bibliography{dnn_bib}




\end{document}


%
\title{Supplentary Document - Unravelling Robustness of Deep Learning based Face Recognition Against Adversarial Attacks}
\maketitle

\begin{figure}[!t]
\centering
\subfloat[Grids]{\includegraphics[width=0.20\linewidth]{./gorig/conv3_2}}\hspace{1em}
\subfloat[Zoomed]{\includegraphics[width=0.20\linewidth]{./gorig/conv3_2_zoom}}\hspace{1em}
\subfloat[Beard]{\includegraphics[width=0.20\linewidth]{./beardorig/conv3_2}}\hspace{1em}
\subfloat[Zoomed]{\includegraphics[width=0.20\linewidth]{./beardorig/conv3_2_zoom}}
\\
\subfloat[Grids]{\includegraphics[width=0.20\linewidth]{./gorig/pool3}}\hspace{1em}
\subfloat[Zoomed]{\includegraphics[width=0.20\linewidth]{./gorig/pool3_zoom}}\hspace{1em}
\subfloat[Beard]{\includegraphics[width=0.20\linewidth]{./beardorig/pool3}}\hspace{1em}
\subfloat[Zoomed]{\includegraphics[width=0.20\linewidth]{./beardorig/pool3_zoom}}
\\
\subfloat[Grids]{\includegraphics[width=0.20\linewidth]{./grids/conv3_2}}\hspace{1em}
\subfloat[Zoomed]{\includegraphics[width=0.20\linewidth]{./grids/conv3_2_zoom}}\hspace{1em}
\subfloat[Beard]{\includegraphics[width=0.20\linewidth]{./beard/conv3_2}}\hspace{1em}
\subfloat[Zoomed]{\includegraphics[width=0.20\linewidth]{./beard/conv3_2_zoom}}
\\
\subfloat[Grids]{\includegraphics[width=0.20\linewidth]{./grids/pool3}}\hspace{1em}
\subfloat[Zoomed]{\includegraphics[width=0.20\linewidth]{./grids/pool3_zoom}}\hspace{1em}
\subfloat[Beard]{\includegraphics[width=0.20\linewidth]{./beard/pool3}}\hspace{1em}
\subfloat[Zoomed]{\includegraphics[width=0.20\linewidth]{./beard/pool3_zoom}}
\caption{\label{fig:visualization} Visualizing filter responses for selected layers from the VGG network when the input image is unaltered and affected by the grids and beard distortions. The first two rows present visualizations for conv3\_2 and pool3 layers for the original input images respectively. The next two rows present visualizations for the same layers when the input images are distorted using adversarial perturbations. The propagation of the adversarial signal into the intermediate layer representations is the inspiration for our proposed detection and mitigation methodologies.}
\end{figure}



